\documentclass{article}
\PassOptionsToPackage{numbers, compress}{natbib}

\usepackage[preprint]{neurips_2021}

\usepackage{wrapfig}
\usepackage{hyperref}
\usepackage{microtype}
\usepackage{graphicx}
\usepackage{subfigure}
\usepackage{amsmath}
\usepackage{bm}
\usepackage{times}
\usepackage{helvet}
\usepackage{courier}
\usepackage{caption}
\usepackage{subfigure}
\usepackage{amsthm,amsmath,amssymb}
\usepackage{mathrsfs}
\usepackage{bbm}
\usepackage{algorithmicx,algorithm}
\usepackage{float}

\usepackage[utf8]{inputenc} 
\usepackage[T1]{fontenc}    
\usepackage{hyperref}       
\usepackage{url}            
\usepackage{booktabs}       
\usepackage{amsfonts}       
\usepackage{nicefrac}       
\usepackage{microtype}      
\usepackage{xcolor}         

\title{Cooperative Multi-Agent Transfer Learning with Level-Adaptive Credit Assignment}

\author {
    Tianze Zhou\textsuperscript{\rm 1}\footnotemark[1] \quad
    Fubiao Zhang\textsuperscript{\rm 1} \quad
    Kun Shao\textsuperscript{\rm 2} \quad
    Kai Li\textsuperscript{\rm 3} \quad
    Wenhan Huang\textsuperscript{\rm 3}
    \AND
    Jun Luo\textsuperscript{\rm 2} \quad
    Weixun Wang\textsuperscript{\rm 4} \quad
    Yaodong Yang\textsuperscript{\rm 2} \quad
    Hangyu Mao\textsuperscript{\rm 2}
    \AND
    Bin Wang\textsuperscript{\rm 2} \quad
    Dong Li\textsuperscript{\rm 2} \quad
    Wulong Liu\textsuperscript{\rm 2} \quad
    Jianye Hao \textsuperscript{\rm 2} \\
    \\
    \\
    $^1$ Beijing Institute of Technology\\
	\{tianzezhou, fubiao.zhang\}@bit.edu.cn\\
	$^2$ Noah's Ark Lab, Huawei Technologoies\\
	\{shaokun2, haojianye\}@huawei.com\\
	$^3$ Shanghai Jiao Tong University\\
	$^4$ Tianjin University \\
}

\begin{document}

\maketitle

\renewcommand{\thefootnote}{\fnsymbol{footnote}}
\footnotetext[1]{Work done during an internship at Noah's Ark Lab, Huawei Technologoies.}



\begin{abstract}
Extending transfer learning to cooperative multi-agent reinforcement learning (MARL) has recently received much attention. In contrast to the single-agent setting, the coordination indispensable in cooperative MARL  constrains each agent's policy. However, existing transfer methods focus exclusively on agent policy and ignores coordination knowledge. We propose a new architecture that realizes robust coordination knowledge transfer through appropriate decomposition of the overall coordination into several coordination patterns. We use a novel mixing network named level-adaptive QTransformer (LA-QTransformer) to realize agent coordination that considers credit assignment, with appropriate coordination patterns for different agents realized by a novel level-adaptive Transformer (LA-Transformer) dedicated to the transfer of coordination knowledge. In addition, we use a novel agent network named Population Invariant agent with Transformer (PIT) to realize the coordination transfer in more varieties of scenarios. Extensive experiments in StarCraft II micro-management show that LA-QTransformer together with PIT achieves superior performance compared with state-of-the-art baselines. 

\end{abstract}

\section{Introduction}

Coordination in multi-agent reinforcement learning (MARL) is a popular topic in fields ranging from robotics \cite{haarnoja2018soft, haarnoja2018learning}, computer games \cite{openai2019dota, ye2020mastering} to recommendation systems \cite{zhao2017deep}. 
Centralized training with decentralized execution (CTDE) is a popular regime in cooperative MARL to realize efficient agent coordination. 
Existing CTDE research covers important topics such as division of agents \cite{wang2020roma}, diversification \cite{yang2020multi} and exploration \cite{mahajan2019maven}. 
Recent works \cite{wang2019more, updet, agarwal2019learning, long2020evolutionary, ijcai2019-65} have also started to make progress in \textit{transfer learning} in cooperative MARL.
For example, \citet{ijcai2019-65} use policy distillation \cite{rusu2015policy} to achieve fixed agent transfer learning. 
However, the agent population varies in different tasks in most cases.
To solve this problem, DyAN \cite{wang2019more} uses a graph neural network to adapt to dynamic agent population. 
UPDeT \cite{updet} uses Transformer\cite{NIPS2017_3f5ee243} to realize a universal and transferable agent policy network to achieve agent-level knowledge transfer.
However, these methods all focus on the transfer of individual agent policy and ignore the coordination knowledge.
Unlike single-agent tasks, cooperative multi-agent tasks require the coordination of multiple agents. 
Ignoring coordination knowledge may lead to biased transfer because the difference in coordination implies a difference in agent policy.

\begin{wrapfigure}{r}{5cm}
\centering
\includegraphics[width=0.4\textwidth]{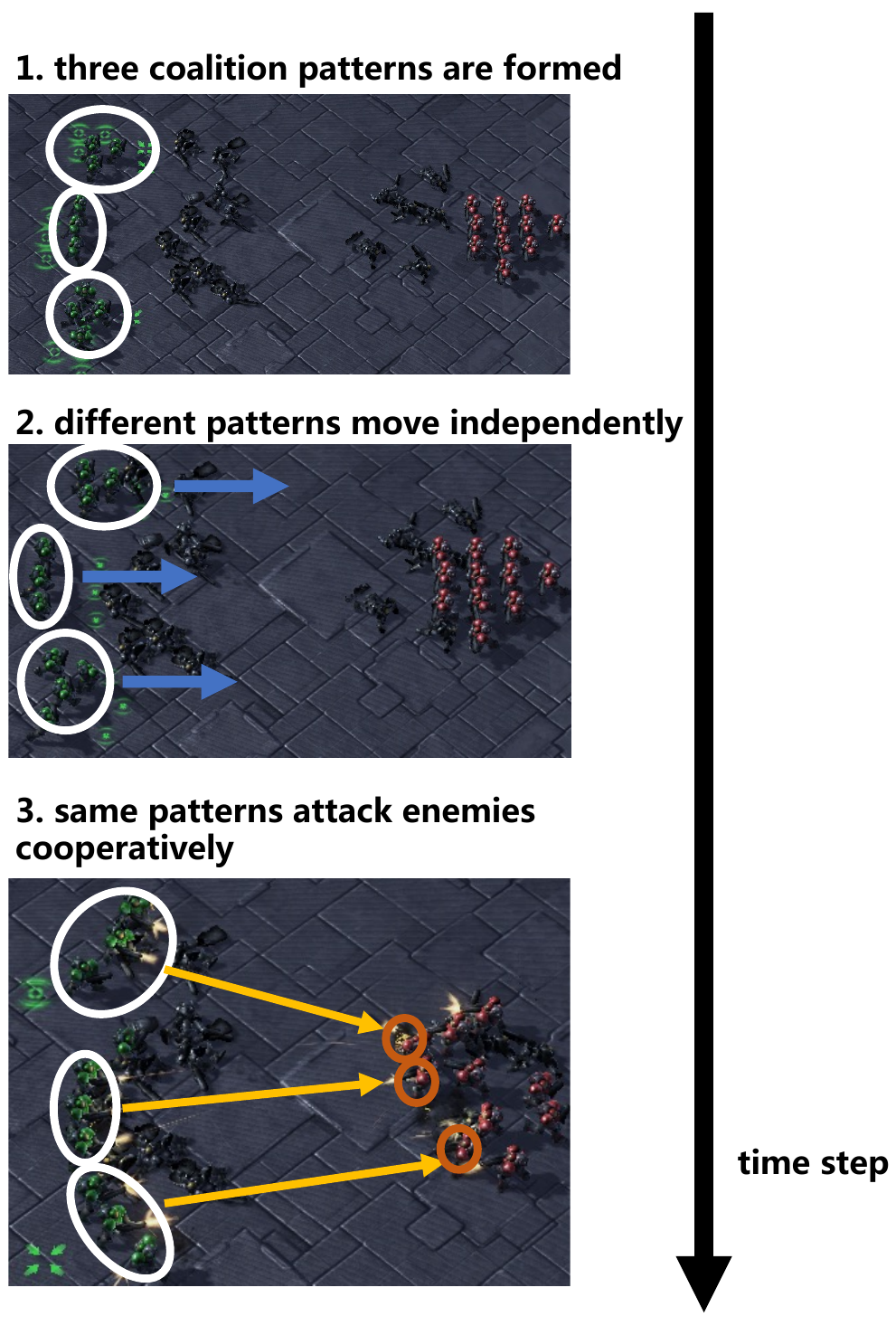}
\caption{Coordination patterns in StarCraft II.}
\label{demo}
\end{wrapfigure}

In cooperative MARL, while the joint policy differs from task to task, the underlying coordination may be decomposed into several patterns that remain valid across different tasks. 
Figure \ref{demo} illustrates this point in StarCraft II, where agents tend to form three different coalition patterns \cite{horling2004survey} with different coalition patterns accomplishing different sub-tasks. 
By leveraging decomposition according to coordination patterns, we may achieve robust coordination knowledge transfer.
The example in Figure \ref{demo} also suggests that the coordination patterns tend to involve a regular number of agents, such as pairwise coordination patterns, triplet coordination, etc.
This means knowledge transfer on coordination policy is manageable in terms of scale.

Transformer \cite{NIPS2017_3f5ee243} is a popular module to capture the relationship among elements and is widely used in nature language process \cite{reformer} and computer vision \cite{ViT}.
In this paper, we use Transformer to capture the correlation between agents and construct the coordination patterns.
However, the traditional Transformer module can only construct the pairwise coordination pattern.
While stacking multiple Transformer modules could allow us to go beyond the pairwise pattern, this approach is not suitable for large-scale multi-agent scenarios due to the enormous computational cost.
Instead, we propose the level-adaptive Transformer (LA-Transformer), which can adaptively capture agent-specific coordination levels and realize coordination patterns involving a variable number of agents. 
We realize the LA-Transformer using both hard attention and a hybrid-based method.
The hard LA-Transformer focuses on the most appropriate coordination level, while the hybrid LA-Transformer merges the features from multiple levels.

Proper credit assignment is essential for coordination among multiple agents in both policy-based \cite{coma, dop} and value-based \cite{VDN} cooperative MARL.  
The credit each agent receives must reflect their contribution towards the coordinated performance.
In CTDE for MARL, a trainable mixing network is often used to implement the required credit assignment.
We follow this practice and introduce LA-Transformer into the design of a novel mixing network named level-adaptive QTransformer (LA-QTransformer).
Compared with other mixing networks such as QMIX \cite{QMIX} and Qatten \cite{Qatten}, LA-QTransformer is a more expressive mechanism for coordination policy learning and coordination knowledge transfer. 

However, redesigning the mixing network alone cannot achieve adequate coordination transfer because the dynamic agent population size limits the agent policy to be reused. 
Previous methods implicitly assume that the joint action space has a fixed dimension or only small and medium-sized scenarios need to be handled. 
Such assumptions make it hard to apply them in real scenarios.
To handle this problem, we design a novel agent structure called Population Invariant agent with Transformer (PIT) to realize generalized coordination knowledge transfer in scenarios of different scales with variable agent numbers.

Evaluation of our new method with the SMAC benchmark \cite{SMAC} shows that it outperforms current SOTA methods in transfer scenarios as well as non-transfer scenarios.
In addition, a curricular training experiment with an increasing number of agents validated the robustness of our method.
Finally, we demonstrate the interpretability of the proposed modules and confirm the contribution of LA-Transformer with ablation studies.

\section{Backgrounds}

\subsection{Cooperative multi-agent Q-learning}

The fully cooperative MARL task can be formulated as a Dec-POMDP \cite{ctde1}. A tuple can represent Dec-POMDP $\left< I, S, U, Z, P, R, O, n,\gamma\right>$, where $s \in S$ represents the global state of the environment. At any time, each agent $i \in I \equiv \{1,...,n \}$ interacts with the environment by generating corresponding action $u_i \in{U}$ through it's local observation vector $z_i \in Z$ according to the observation function $O(s, i)$. The overall objective is to maximize the cumulative reward $R$ from environmental feedback. The environment receives the joint action $\mathbf{a}$, and transfers to the next state $\mathbf{s^{\prime}}$ according to the state transition function $P\left(s^{\prime} \mid s, \mathbf{a}\right)$. $n$ defines the number of agents, and $\gamma$ represents the discount factor. 

Centralized training with decentralized execution  \cite{ctde2} (CTDE) is a popular regime to address the Dec-POMDP problem.
In the CTDE framework, the mixing network is introduced to merge all individual Q values into $Q_{tot}$:
\begin{align}
Q_{tot}(\mathbf{\tau}, \mathbf{u}, s ; \theta) = f([Q_i(\tau^i, u^i)]_i^n, s; \theta).
\end{align}
And then TD-learning is used to train the whole network
\begin{align}
\mathcal{L}(\theta) = \sum_{i=1}^b \left[\left(y_{i}^{t o t}-Q_{t o t}(\mathbf{\tau}, \mathbf{u}, s ; \theta)\right)^{2}\right],
\end{align}
where $b$ is the batch size of replay buffer, $y^{\text {tot }}=r+\gamma \max _{\mathbf{u}^{\prime}} Q_{\text {tot }}\left(\tau^{\prime}, \mathbf{u}^{\prime}, s^{\prime} ; \theta^{-}\right)$, and $\theta^{-}$ is the parameter of target network.

\subsection{Multi-agent transfer learning}

The basic idea behind transfer learning is that the knowledge acquired from previous tasks can be reused to accelerate learning drastically, and it makes the learning of complex tasks feasible \cite{da2019survey, boutsioukis2011transfer}. Due to the complexity of MARL, multi-agent transfer learning is not a straightforward extension of single-agent transfer learning. In the multi-agent setting, the policy mapping expands from a single agent to multiple agents, and the dimension of the mapping is varying with specific tasks:
\begin{align}
J_{p} \rightarrow J_{c}: A_{1} \times \ldots \times A_{n} \rightarrow A_{1}^{\prime} \times \ldots \times A_{m}^{\prime}
\end{align}
where $J_{p}$ and $J_{c}$ represent the joint-policy in the previous task and the current task, respectively, and $n, m$ shows the number of agents in these tasks.

\cite{da2019survey} divides multi-agent transfer learning (MATL) into two main types: the intra-agent transfer and the inter-agent transfer. The intra-agent transfer focuses on the relationship between the source tasks and the target tasks, while the inter-agent transfer pays more attention to reusing knowledge received from communication with other agents. 

\subsection{Transformer}

The Transformer is an attention-based neural network structure widely used in nature language process and computer vision. The traditional Transformer module consists of two sub-structures, the attention module and the feed-forward network. Soft attention and hard attention are two approaches to realize the attention mechanism.
\textit{Soft attention} takes the softmax function to calculate the input elements relationship.
\begin{align}
	\text { Attention }(\mathbf{Q}, \mathbf{K}, \mathbf{V})=\operatorname{softmax}\left(\frac{\mathbf{Q} \mathbf{K}^{T}}{\sqrt{d_{k}}}\right) \mathbf{V},
\end{align}
where $\mathbf{Q}, \mathbf{K}, \mathbf{V}$ represents the query, keys, values of input elements respectively and $\sqrt{d_{k}}$ is the normalization coefficient.

Due to element weights is calculated directly, soft attention is fully differentiable. However, the softmax function weakens the ground truth element's weight, limiting the actual performance. \textit{Hard attention} overcomes the limitation of soft attention by selecting the sole element. However, this selecting operation is non-differentiable. Gumbel softmax \cite{jang2016categorical} is a popular trick to approximate hard attention performance while keeping the neural network back-propagation differentiable. To enhance the representation of the embedding features, Transformer utilizes a feed-forward network. The feed-forward network contains a 1-D convolutional layer and a layer-normalization module.

Transformer is suitable to capture elements relationship in cooperative MARL. Due to Transformer's flexible I/O characteristic, it can handle dynamic element inputs. Compared with RNN-based methods, Transformer does not care about the order of elements and can process elements in parallel.

\section{Methods}

\begin{figure*}[htb]
	\centering
	\includegraphics[width=1.0\textwidth]{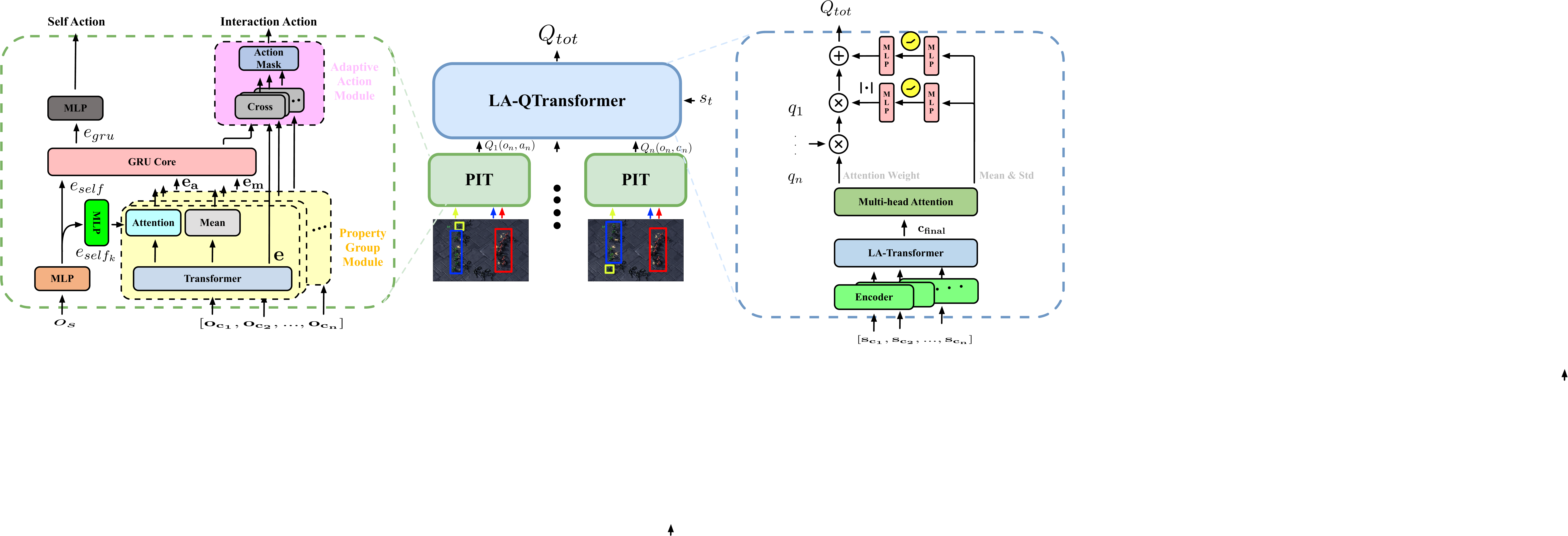}
	\caption{\textbf{Left:} The structure of PIT. Firstly, The observation separated explicitly is flowed into the property group module to generate the different class embedding features. The GRU core is utilized to merge all embedding features. Finally, the adaptive action module utilizes the GRU embedding feature and the property group module embedding features to generate dynamic agent actions. \textbf{Middle:} The whole structure of our methods. \textbf{Right:} The structure of LA-QTransformer. LA-QTransformer first separates the state features into different class entities' features and utilizes the encoder layers to encode the features into the same dimension. Then the LA-Transformer module is used to generate multi-level coordination patterns and merges(selects) the appropriate coordination patterns. The multi-head attention module then integrates the coordination patterns and generates agent credit values.}
	\label{total}
\end{figure*}

In this section, we design a novel value-based framework to realize coordination knowledge transfer in cooperative MARL. Figure \ref{total} describes the whole structure of our methods. It contains a mixing network level-adaptive QTransformer (LA-QTransformer) that utilizes the level-adaptive Transformer (LA-Transformer) module to realize the coordination knowledge transfer and the agent network, Population Invariant agent with Transformer (PIT), to achieve coordination transfer in universal scenarios. 

\subsection{Level-Adaptive QTransformer} 

\subsubsection{Level-Adaptive Transformer}

\begin{figure}
\centering
\includegraphics[width=0.4\textwidth]{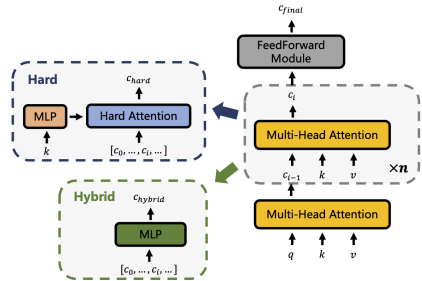}
\caption{Two implementations of LA-Transformer.}
\label{ms-Transformer}
\end{figure}

The Transformer module can be applied to generate the pairwise coordination patterns via capturing the relationship of input elements. However, only considering the pairwise coordination patterns can not achieve general coordination transfer. To generate coordination patterns on multiple levels, a native method is stacking Transformer modules. However, this is unrealistic in large-scale multi-agent scenarios due to its massive memory consumption. Besides, stacking modules blurs different level relationships and makes it impossible to determine which level works. In this section, we propose an efficient module named level-adaptive Transformer (LA-Transformer) to generate multi-level coordination patterns and capture the most suitable coordination pattern. We implement LA-Transformer with two methods, including hard attention and the hybrid method, as shown in Figure \ref{ms-Transformer}.

In the traditional Transformer, the pairwise coordination patterns are generated via: 
\begin{align}
	\mathbf{Q, K, V} = \operatorname{MLP}_{Q,K,V}(\bold{s}),  \\
	c =\operatorname{Softmax}\left(\frac{\mathbf{Q} \mathbf{K}^{T}}{\sqrt{d_{k}}}\right) \mathbf{V}, 
\end{align}
where $c$ is the coordination pattern. $\frac{\mathbf{Q} \mathbf{K}^{T}}{\sqrt{d_{k}}}$ can be approximated as the coordination relationship.
Compared with traditional stacking operations, we fix $k$ and $v$ to realize linearly increasing coordination level and prevent feature blurring, respectively. The specific coordination level is calculated via:
\begin{align}
	c_i &=\operatorname{Softmax}\left(\frac{\mathbf{c_{i-1}} \mathbf{K}^{T}}{\sqrt{d_{k}}}\right) \mathbf{V},  
\end{align}
where $c_{i-1}$ is the coordination pattern of previous level.

Furthermore, we analyze that the upper bound of the coordination pattern level does not need to be huge. The relationship mapping of the adjacent levels tends to be stationary as the level increases according to the Brouwer Fixpoint Theorem \cite{kellogg1976constructive}. With the coordination level increasing, there exist two coordination patterns with adjacent level $c_i, c_j$ that their difference tends to zero: 
\begin{align}
	\left\| c_i-c_j\right\|_2 \leq \epsilon, \ \exists i, j, \; i \geq j \geq 0, 
\end{align}

Due to the difference among agents, we propose two methods to generate different levels of coordination patterns for different agents.

\textbf{LA-Transformer (hard)}. LA-Transformer (hard) utilizes the hard attention mechanism to select the most appropriate coordination patterns.
To achieve the hard attention-based LA-Transformer, we first utilize the MLP function to encode the initial embedding features to get the key of the hard attention, i.e., $\mathbf{k_e}$, and then take the Gumbel softmax function to generate the mask on different levels.
\begin{align}
	\mathbf{k_e} &= \operatorname{MLP}_{K}(\bold{s}),  \\
	\mathbf{mask} &= \operatorname{gumbel\_softmax}(k_e [c_1, ... c_i])
\end{align}
Then, we take the mask to select the agent-specific coordination pattern. 
 \begin{align}
	\mathbf{c_{hard}} = \mathbf{mask} \times [c_1, ..., c_i]
\end{align}
The advantage of utilizing the mask to select is that it can explicitly provide the value of the level.

\textbf{LA-Transformer (hybrid)}. Although the LA-Transformer (hard) can explicitly select a coordination pattern, it inevitably ignores some essential details. So we propose the hybrid level-adaptive Transformer, which generates the level-adaptive coordination pattern via adaptively fusing coordination patterns from all levels.
Specifically, we take the MLP function to do the fuse operation.
\begin{align}
	\mathbf{c_{hybrid}} &= \operatorname{MLP}([c_1, ..., c_i])
\end{align}

Finally, the coordination patterns $\mathbf{c_{hard}}$ or $\mathbf{c_{hybrid}}$ flows into a FeedForward Module to enhance the representation ability and generate $\mathbf{c_{final}}$.


\subsubsection{Level-Adaptive Transformer-based mixing network (LA-QTransformer)}

Limited by the dueling structure mixing network (such as QPLEX) large search space, we take the popular QMIX-like monotonic mixing network as the baseline model. Our framework can be expressed as $Q_{tot}=\sum_i^m w_i(s)q_i(o_i, \tau_i)$, where $w_i(s)$ is the non-negative parameter realized by the proposed mixing network (LA-QTransformer), and $q_i(o_i, \tau_i)$ is the individual Q value of agent $i$.

The right part of figure \ref{total} shows the structure of LA-QTransformer. LA-QTransformer has two essential modules, the coordination decomposition module and the coordination integration module. In the coordination decomposition module, LA-QTransformer utilizes the LA-Transformer module to generate different coordination patterns for different agents. After that, LA-QTransformer utilizes a multi-head attention module to combine all agents' coordination patterns and generate agents' credits.

Due to the input entity's different classes and dimensions, we first divide the state features into several entities via the entity class and utilize the MLP-based encoder to embed all entities' features into the same dimension. Then we take the preprocessed state features into the LA-Transformer module to generate suitable coordination patterns. After that, we utilize the multi-head attention module to combine all coordination patterns and generate the credit assignment weights. Finally, LA-QTransformer takes the dot product operation to merge agents' Q values and generate the total Q value $Q_{total}$. In addition, a bias term is used to make up for the residual.




\subsection{Population invariant agent via Transformer (PIT)}

To realize the coordination transfer in more general scenarios, we design the Population Invariant agent via Transformer (PIT), as shown in the left of Figure \ref{total}. PIT has three main parts, the \textit{Property Group Module}, the \textit{GRU Core}, and the \textit{Adaptive Action Module}. In the Property Group Module, PIT explicitly groups the input entities via entities' property and generates group embeddings. The GRU Core merges all group embeddings. The Adaptive Action Module makes it adaptive to the dynamic action space from different scenarios.

\textit{Property Group Module}. We first divide the observation into the agent attribute features $o_{self}$, and several group feature sets $\mathbf{o_{special}}$ via the entity's property. For example, in SMAC, the role property, such as ally and enemy, can be used to divide groups. Due to solid relevance in the same groups, we introduce the Transformer module to generate adaptable and general relevant embeddings. To solve the dynamic entity population problem, we unify all entities in the same groups and represent these features on the group level. Inspired by the mean-field method and attention mechanism, we represent the group features with the mean features of all inner group entities $e_m$ and the most relevant features $e_i$.

\textit{GRU Core}. The GRU Core utilizes the agent-self features $e_{self}$ and different group representation $e_{m, i}, e_{a, i}$ to capture the temporal change of the group feature and merge all of these features.

\textit{Adaptive Action Module}. Inspired by the Action Semantics Network \cite{wang2019action}, we classify actions to adapt to dynamic action space. We divide the total action space into self-related actions, such as move and no-op, and interacting actions, such as attack. Considering that the interacting action is highly related to interacting entities, we utilize the embedding features generated from Transformer to construct interacting action. Note that some properties do not include the action attribute, so we add an action mask module to block unrelated actions. 

The details of PIT can be seen in Appendix.

\section{Experimental results}

\begin{figure*}[htb]
\centering
\includegraphics[width=1.0\textwidth]{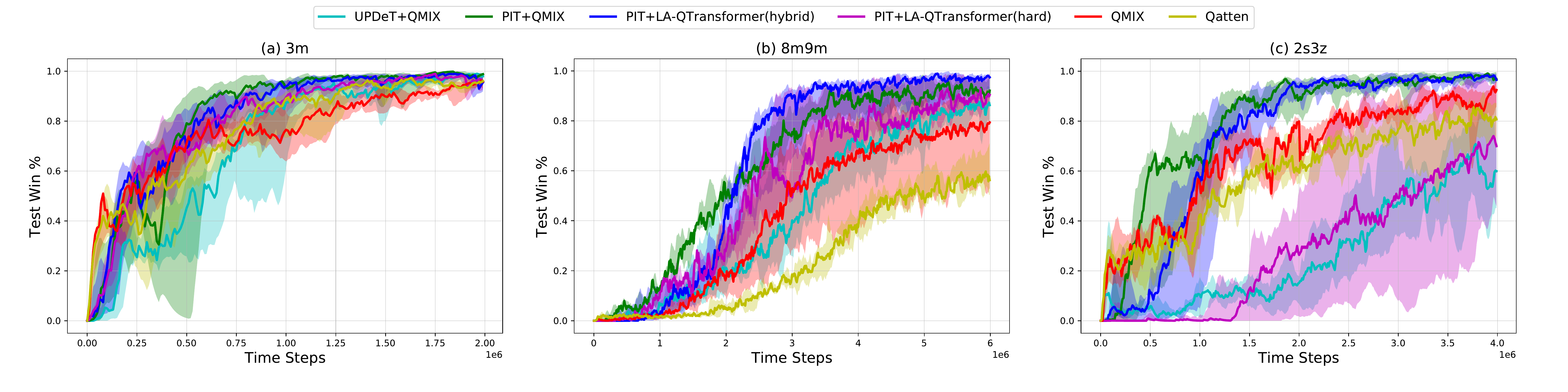}
\caption{The performance of PIT LA-QTransformer and PIT  in SMAC. 
We compare our methods with several baselines in homogeneous and heterogeneous scenarios.
}
\label{baseline_result}
\end{figure*}

We evaluate the performance of PIT and LA-QTransformer in SMAC, which is a popular MARL benchmark based on the real-time strategy game StarCraft II. In SMAC, each unit is controlled by an independent agent with local observation. In contrast, the opponent’s units are controlled by the built-in rule-based AI. To test the robustness of our methods, all experiments are run with five random seeds and evaluated under seven threads in parallel. 

\subsection{Baseline performance} 

Figure \ref{baseline_result} shows the performance in small-sized homogeneous, large-sized homogeneous, and median-size heterogeneous scenarios, respectively. In the simple 3m scenario, both LA-QTransformer(hybrid) and LA-QTransformer(hard) achieve excellent performance. In more challenging scenarios, LA-QTransformer(hybrid) outperforms other baseline methods. However, due to missing details of selecting one specific coordination pattern, LA-QTransformer(hard) method shows suboptimal performance.

\subsection{Performance in transfer learning setting}

\begin{figure*}[htb]
\centering
\includegraphics[width=1.0\textwidth]{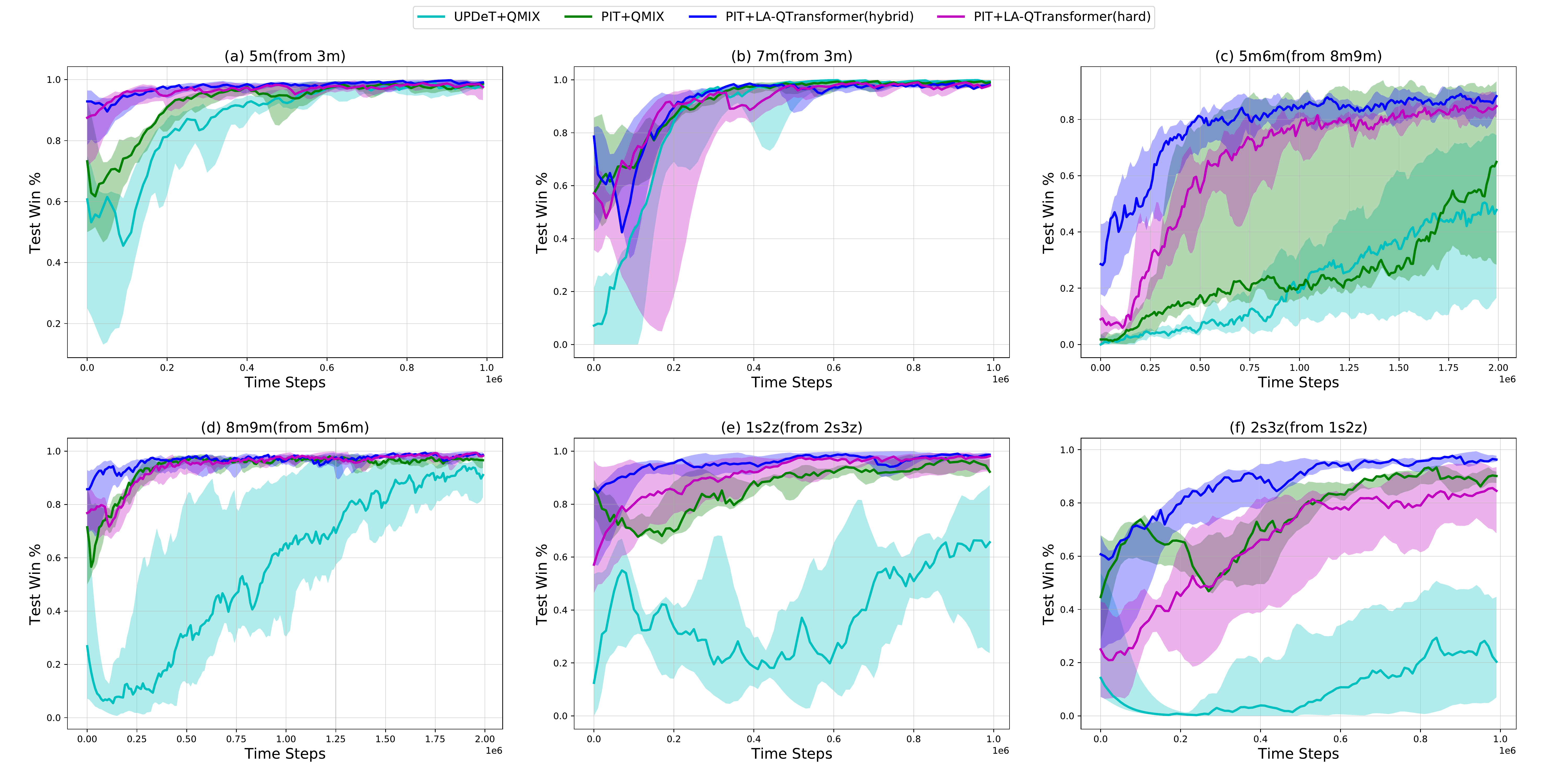}
\caption{The experimental results of PIT and UPDeT with mixing modules, LA-QTransformer and QMIX. Agents are well-trained to converge in the original scenarios and then fine-tuned in the latter scenarios.}
\label{transfers}
\end{figure*}

Figure \ref{transfers} shows the evaluated performance of PIT with LA-Transformer and QMIX and current state-of-the-arts multi-agent transfer learning method UPDeT. The experimental results present the performance of different modules in the transfer learning setting.

\textbf{Transfer in different scales}. Figure \ref{transfers}(a) shows that both two LA-QTransformer methods can achieve excellent coordination knowledge transfer while QMIX is unstable. In Figure \ref{transfers}(b), we conduct a more extensive transfer test in different scales (more than two times). Due to the enormous scale changes, LA-QTransformer needs to regenerate proper coordination patterns to adapt to the difference of scenarios, which shows little instability at the beginning of the training process. However, LA-QTransformer has strong adaptability and can quickly converge to the optimal policy. Besides, the experimental result demonstrates that PIT is superior to UPDeT.

\textbf{Transfer in different difficulty levels}. We evaluate the performance of coordination knowledge transfer in two scenarios with different difficulty levels (8m\_vs\_9m is simple and 5m\_vs\_6m is complex). Figure \ref{transfers}(c) and \ref{transfers}(d) show the transfer from simple to complex and from complex to simple respectively. LA-QTransformer shows an advantage in the jumpstart (the initial performance) and the asymptotic performance. Due to no coordination knowledge transfer, QMIX's performance is unstable in transferring from simple to complex.

\textbf{Transfer in heterogeneous scenarios}. As shown in Figure \ref{transfers}(e) and \ref{transfers}(f), we test the performance in two heterogeneous scenarios with different scales. The LA-Transformer(hybrid) outperforms other baselines because the coordination patterns in heterogeneous scenarios are also helpful. In heterogeneous scenarios, the coordination patterns in different types of agents have obvious differences, making LA-Transformer(hard) more likely to miss valuable information. This results in suboptimal performance. Besides, the empirical results show that UPDeT does not perform well in heterogeneous scenarios. 

\subsection{Performance with curricular transfer learning}

\begin{figure*}[htb]
\centering
\includegraphics[width=1.0\textwidth]{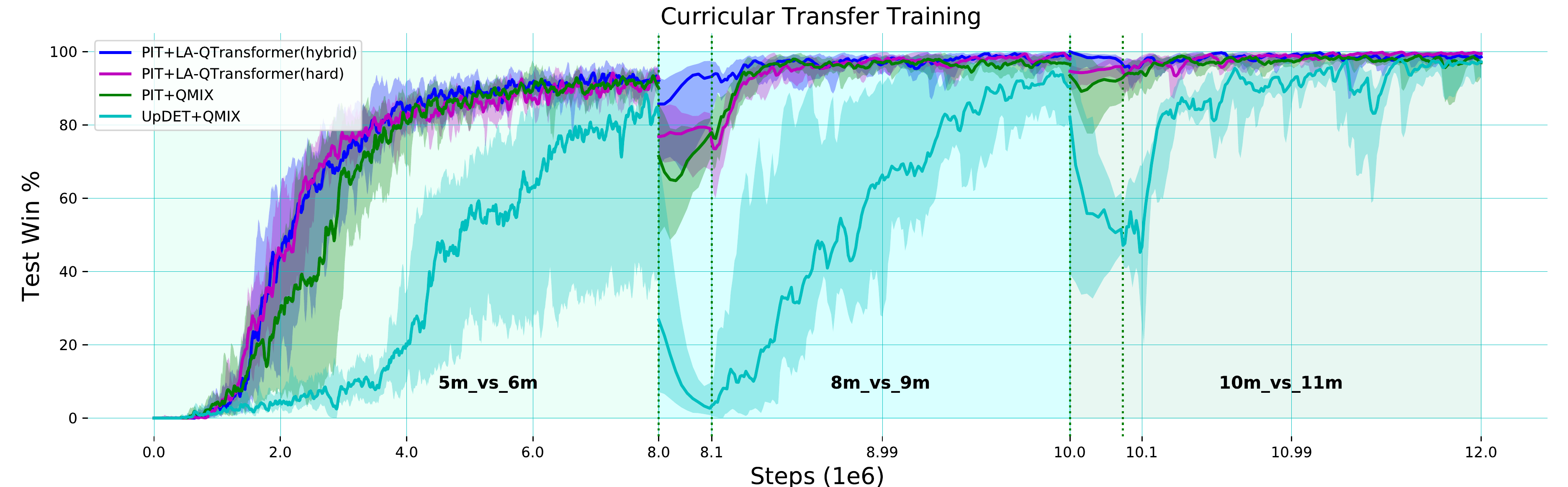}
\caption{The curricular transfer experiment in SMAC scenarios.}
\label{phrase_training}
\end{figure*}

We extend our methods to curricular learning and make the scaling up experiment. Figure \ref{phrase_training} shows that agents are firstly trained in the 5m\_vs\_6m scenario and then transferred to the 8m\_vs\_9m scenario with 2M training steps, and finally tested in the 10m\_vs\_11m scenario. According to the results, curricular learning can correct coordination patterns and generate more general coordination patterns. The performance in the 10m\_vs\_11m scenario shows that LA-QTransformer with PIT achieves excellent performance, even without any further training.  

\subsection{Interpretation of LA-QTransformer}

\begin{figure*}
\centering
\subfigure[Credit assignment values in two different scenarios]
{
	\begin{minipage}{6.5cm}
	\centering          
	\includegraphics[width=1.0\textwidth]{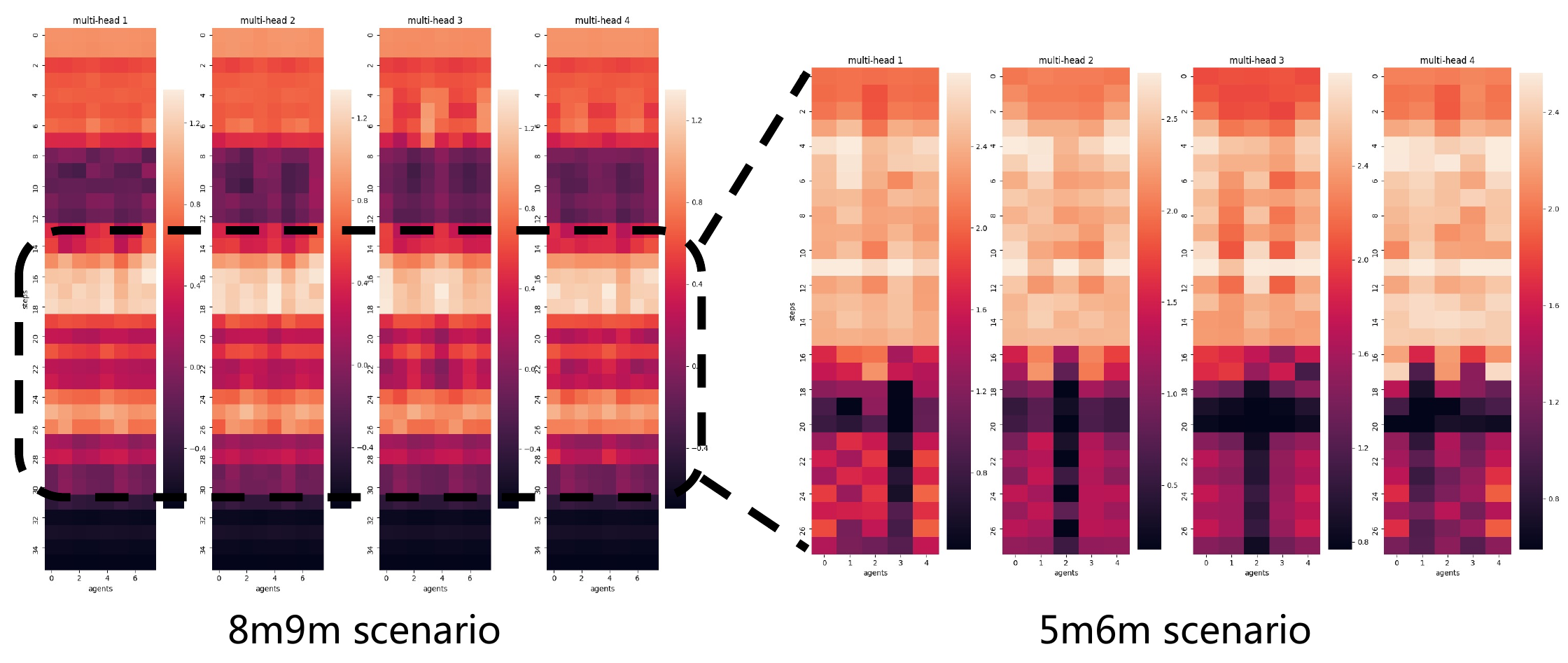}   
	\end{minipage}
}
\subfigure[Weight of pairwise-coordination.] 
{
	\begin{minipage}{6.5cm}
	\centering      
\includegraphics[width=1.0\textwidth]{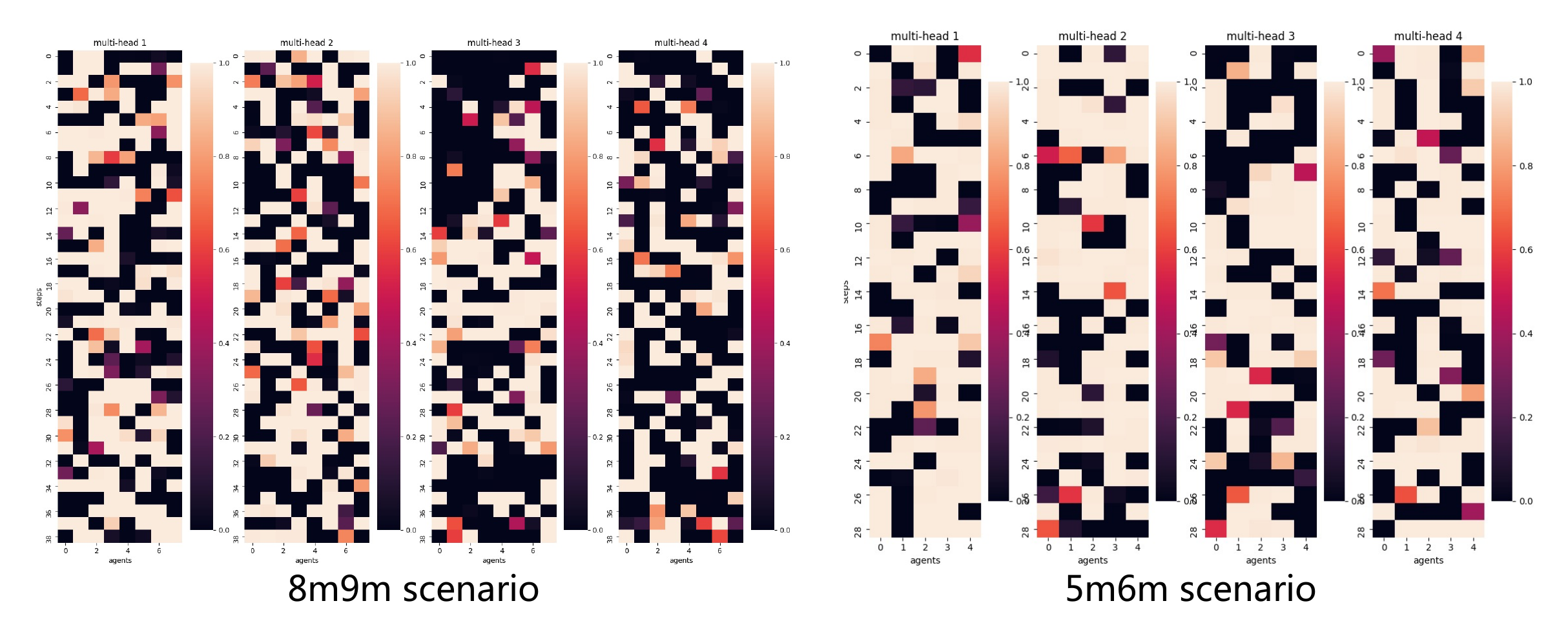}   
	\end{minipage}
}
\caption{The interpretation of LA-QTransformer.}
\label{analysis}  
\end{figure*}


\textbf{Transfer ability}. Figure \ref{analysis} (a) shows the temporal credit assignment values of LA-QTransformer, and it verifies the feasibility of coordination transfer. The Figure shows that the second half credit assignment values of the 8m\_vs\_9m scenario have a remarkable similarity with that in the 5m\_vs\_6m scenario, demonstrating the coordination of the coordination policies. 


\textbf{Coordination level of LA-QTransformer}. In Figure \ref{analysis} (b), we present the pairwise coordination weight in LA-QTransformer and find that the complex scenario needs higher-level coordination. Because the average pairwise coordination weight of 5m\_vs\_6m scenario is smaller than that in the 8m\_vs\_9m scenario. This also indicates that learning the whole coordination policy has difficulty in realizing the coordination knowledge transfer due to policy overfitting.

\textbf{Initial win rate analysis}. Table \ref{testwinrate} emphasizes the jumpstart performance of different mixing networks under the PIT. The jumpstart metric can measure the agent performance without training and shows method generalization. Compared with QMIX, LA-QTransformer can learn a more robust coordination policy via coordination decomposition. 

\begin{table}[htbp]
	\centering
	\caption{Test win rate of PIT without training.}
	\begin{tabular}{cccc}
		\toprule  
	 new (origin) & LA-QTransformer(hybrid) & LA-QTransformer(hard) & QMIX\\
	    \midrule  
	5m (3m)     & \textbf{97.2\%} &         97.1\%  & 95.3\% \\
	7m (3m)     & \textbf{97.2\%} &         96.9\%  & 95.9\% \\
	8m9m (5m6m) &         95.1\%  & \textbf{97.1\%} & 96.1\% \\
	5m6m (8m9m) & \textbf{83.2\%} &         77.0\%  & 21.7\% \\
	1s2z (2s3z) & \textbf{98.1\%} &         97.1\%  & 91.5\% \\
	2s3z (1s2z) & \textbf{90.8\%} &         47.7\%  & 84.8\% \\
	\bottomrule  
	\end{tabular}
	\label{testwinrate}
\end{table}

\subsection{Ablation study on ML-Transformer}

\begin{wrapfigure}{r}{6.5cm}
\centering
\includegraphics[width=0.4\textwidth]{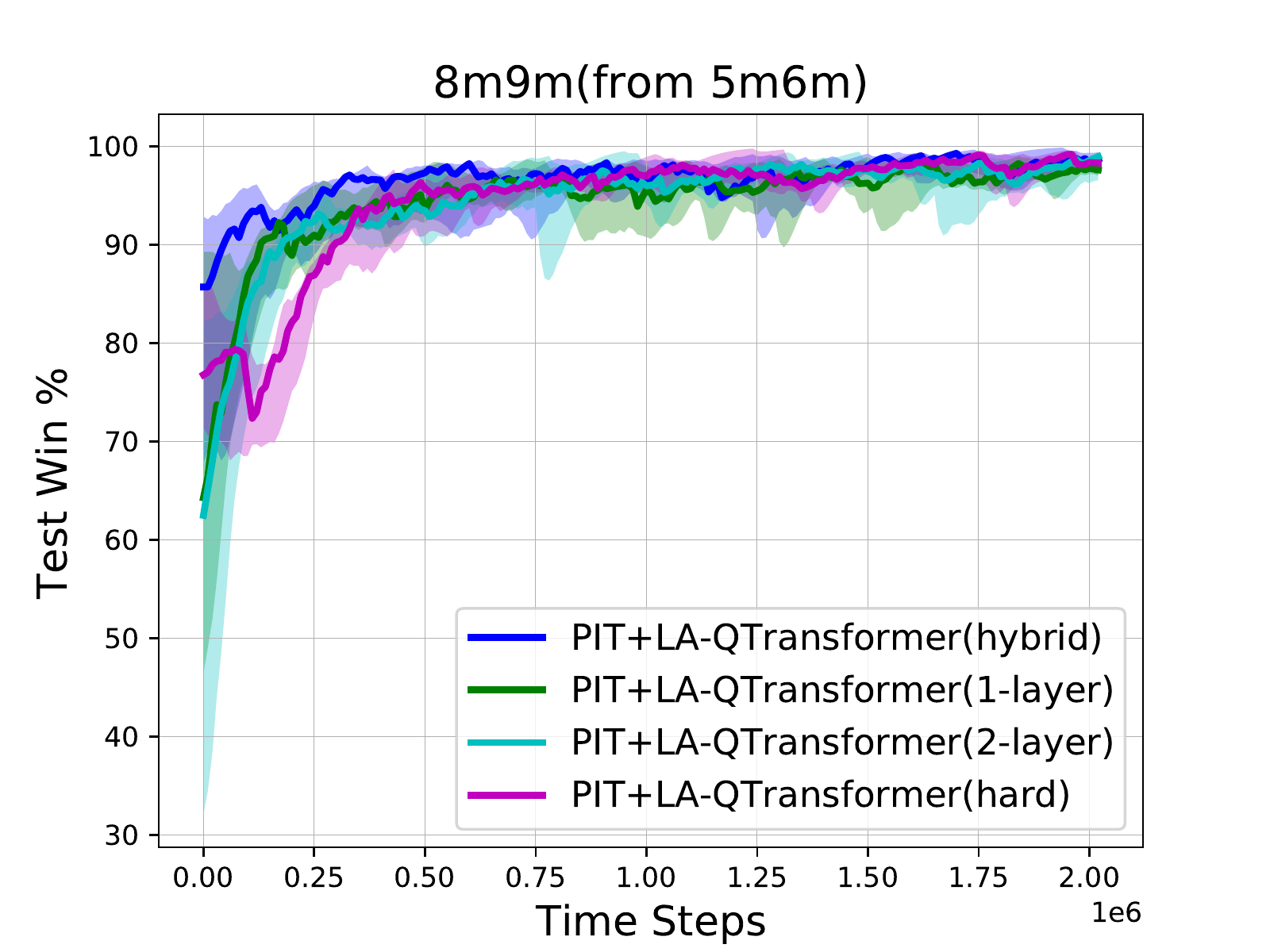}
\caption{Ablation study on the Transformer stacking method.}
\label{ablation}
\end{wrapfigure}

We evaluate the difference between the LA-Transformer and the traditional Transformer stacking method. Figure \ref{ablation} shows that stacking two Transformer layers does not lead to any performance improvement. In cooperative MARL, stacking Transformer modules blurs different level relationships and has difficulty in capturing the proper coordination level. However, LA-Transformer can explicitly distinguish the differences between different coordination levels and perceive the suitable coordination level. LA-Transformer(hard) can be explained as the combination of the two stacking Transformer modules and select the proper coordination level, while LA-Transformer(hybrid) implicitly generates the coordination level. The initial win rate verifies that. 

\section{Related work}
\textbf{Credit assignment methods in cooperative multi-agent Q-learning}. VDN \cite{VDN} uses a simple sum operation to generate the global Q value. QMIX \cite{QMIX}, as an extension of VDN, introduces the monotonicity hypothesis to satisfy the IGM condition and uses the hyper-network to achieve it \cite{ha2016hypernetworks}. Qatten \cite{Qatten} introduces the multi-head attention mechanism to construct the mixing network to obtain better performance. QPLEX \cite{wang2020qplex} uses the dueling network to avoid direct optimization from monotonicity assumptions. Previous methods always focus on the whole coordination policy and can achieve excellent performance in several cooperative tasks. However, learning on the whole coordination level may lead to the over-fitting coordination policy and is unsuitable for transfer learning tasks.  

\textbf{Multi-agent transfer learning methods}. Current MATL has two branches: the auxiliary training technique and the adaptive network structure. Reusing replay buffer and policy distillation are the prevalent auxiliary training methods. \cite{wang2019more} improves the efficiency of value-based MATL by reusing the transition data generated in previous scenarios.  Inspired by policy distillation \cite{rusu2015policy}, \citet{ijcai2019-65} proposes to transfer the knowledge learned in a single agent to multiple agents and uses the n-step return to approximate the difference of the local environment dynamics. This can achieve selective migration and avoid the negative transfer. 

In adaptive network structure methods, agents can directly reload the previous knowledge via adapting to the dynamic observation and action shape. DyAN \cite{wang2019more} uses the graph neural network to decompose the observation into each entity node to deal with the uncertain population of entities. Unlike DyAN, EPC-MADDPG \cite{long2020evolutionary} merges varying entity features to fixed-dimensional features with the attention mechanism. UPDeT \cite{updet} firstly proposes to use Transformer to handle dynamic features. It separates the observation features into several entity-based features and uses the Transformer module to generate different actions. However, previous methods are all limited in realizing the coordination knowledge transfer.



\section{Conclusion}
\label{sec:conclusion}

In this paper, we propose a novel mixing network for cooperative MARL, called LA-QTransformer, to achieve coordination knowledge transfer. 
Compared with the agent-level knowledge transfer, coordination transfer has better generalization and scalability. 
Our network first decomposes the correlations among agents into a series of agent-specific coordination patterns via the level-adaptive Transformer (LA-Transformer) and then integrates the coordination patterns for the purpose of credit assignment.
To ensure the coordination knowledge transfer in more varieties of scenarios, we design a novel agent structure named population invariant agent with Transformer. 
Experiments on the SMAC benchmarks show that LA-QTransformer can achieve excellent coordination policy transfer and outperforms current SOTA baselines.

Through experiments, we notice that curriculum learning can correct the coordination patterns and realize efficient coordination transfer. 
An interesting question then is how to design a systematic curriculum to achieve more efficient coordination transfer. 
Moreover, it should be noted that while we only realized coordination knowledge transfer in multi-agent Q-learning, we may also consider extending our approach to policy-based methods, such as MADDPG \cite{lowe2017multi} and MAPPO \cite{yu2021surprising}.

\bibliographystyle{plainnat}
\bibliography{content/references.bib}

\end{document}